\documentclass[11pt]{article}

\usepackage[final]{acl}

\usepackage{times}
\usepackage{latexsym}

\usepackage[T1]{fontenc}

\usepackage[utf8]{inputenc}

\usepackage{microtype}

\usepackage{inconsolata}

\usepackage{graphicx}

\usepackage{amsmath}
\usepackage{multirow}
\usepackage{booktabs} 
\usepackage{authblk}
\title{LLM-Guided Semantic Bootstrapping for Interpretable Text Classification with Tsetlin Machines}

\begin{document}

\author[1]{Jiechao Gao} 
\author[2]{Rohan Kumar Yadav} 
\author[3]{Yuangang Li} 
\author[1]{Yuandong Pan}
\author[1]{\\Jie Wang} 
\author[4]{Ying Liu} 
\author[1]{Michael Lepech} 

\affil[1]{Stanford University} 
\affil[2]{Independent Researcher} 
\affil[3]{University of California, Irvine} 
\affil[4]{University of the Chinese Academy of Sciences} 


\maketitle
\begin{abstract}
Pretrained language models (PLMs) like BERT provide strong semantic representations but are costly and opaque, while symbolic models such as the Tsetlin Machine (TM) offer transparency but lack semantic generalization. We propose a semantic bootstrapping framework that transfers LLM knowledge into symbolic form, combining interpretability with semantic capacity. Given a class label, an LLM generates sub-intents that guide synthetic data creation through a three-stage curriculum (seed, core, enriched), expanding semantic diversity. A Non-Negated TM (NTM) learns from these examples to extract high-confidence literals as interpretable semantic cues. Injecting these cues into real data enables a TM to align clause logic with LLM-inferred semantics. Our method requires no embeddings or runtime LLM calls, yet equips symbolic models with pretrained semantic priors. Across multiple text classification tasks, it improves interpretability and accuracy over vanilla TM, achieving performance comparable to BERT while remaining fully symbolic and efficient.

\end{abstract}

\section{Introduction}

The Tsetlin Machine (TM) has recently gained traction in artificial intelligence (AI) due to its transparent learning process, interpretable structure, and fully explainable outputs~\cite{Granmo2018TheTM}. It has shown promising results in various natural language processing (NLP) tasks, including document classification~\cite{10.24963/ijcai.2023/378}, sentiment analysis~\cite{Yadav2021HumanLevelIL}, topic classification~\cite{yadav-etal-2021-enhancing}, and fake news detection~\cite{bhattarai-etal-2022-explainable}. Thanks to its clause-based symbolic logic, the TM is particularly suitable for high-stakes domains such as legal and medical document analysis, where interpretability and decision traceability are critical~\cite{Saha2022InterpretableTC,gao2025enhancing,Berge2018UsingTT}.

Despite these advantages, the TM’s symbolic nature imposes limitations. Operating on Boolean bag-of-words (BoW) representations, it struggles to generalize across semantically related terms unless explicitly observed in training. In contrast, pretrained language models (PLMs) such as BERT provide strong semantic representations by capturing contextual relationships~\cite{Devlin2019BERTPO}, but they are costly and opaque. Prior attempts to close this gap—for example, enriching TM inputs with word vectors from embedding models like Word2Vec or GloVe~\cite{yadav-etal-2021-enhancing}—have achieved only limited semantic alignment.

To address this, we propose an \textbf{LLM-guided semantic bootstrapping framework} that integrates high-level semantic knowledge from large language models (LLMs) into the TM pipeline through symbolic augmentation rather than embeddings. The framework prompts an LLM to decompose class labels into interpretable sub-intents (e.g., \textit{positive\_due\_to\_plot}) and generate synthetic training samples via a structured curriculum. These examples guide clause formation in a Non-Negated TM (NTM), whose extracted literals enrich real-world data with interpretable semantic features. A standard TM fine-tuned on this hybrid representation benefits from LLM-inferred semantics while preserving symbolic transparency and efficiency. Our approach thus bridges the gap between symbolic reasoning and pretrained language understanding without requiring embeddings or LLM inference at runtime.

\section{Related Work}

\textbf{Semantic Modeling and Symbolic Interpretability: }Pretrained transformer-based language models such as BERT, T5, and GPT have achieved state-of-the-art performance across a wide range of NLP tasks~\cite{Devlin2019BERTPO,Raffel2019ExploringTL,llm,Liu2019FinetuneBF}. However, their high computational cost and lack of interpretability limit their use in high-stakes domains such as healthcare or law~\cite{Rudin2018StopEB,jain-wallace-2019-attention,wu-etal-2023-transparency}.

Symbolic models offer an interpretable alternative. The Tsetlin Machine (TM), in particular, provides clause-level transparency through propositional logic~\cite{Granmo2018TheTM}, and has been applied to sentiment analysis, topic classification, and explainable fact-checking~\cite{Yadav2021HumanLevelIL,yadav-etal-2021-enhancing,bhattarai-etal-2022-explainable}. However, its reliance on literal bag-of-words representations limits its ability to generalize across semantically similar inputs unless explicitly encoded. Augmenting TM inputs with static embeddings like GloVe~\cite{yadav-etal-2021-enhancing} offers limited semantic alignment and fails to capture context-dependent meaning.

\textbf{LLM-Based Supervision for Interpretable Systems: }Recent approaches have explored leveraging large language models (LLMs) to generate weak supervision signals, synthetic data, or structural priors through prompt-based interaction~\cite{Ding2024DataAU,Hsieh2025DALLMCC,10.1145/3706598.3713491}. These strategies aim to incorporate the semantic capabilities of LLMs without depending on them at inference time. Concurrently, symbolic distillation methods attempt to extract rules or interpretable logic from LLM outputs~\cite{mcdonald-emami-2024-trace,xie-etal-2025-simple,xu-etal-2024-symbol}, though they typically target decision tree structures or linear rules.

Our work builds on these ideas by enabling clause-based symbolic models to inherit semantic priors through structured sub-intent supervision. Unlike prior methods, we maintain full symbolic transparency and efficiency, without embedding layers or runtime LLM calls.


\section{Proposed Model}\label{sec:proposed}

We propose a LLM-guided semantic bootstrapping framework for interpretable text classification with Tsetlin Machines, without relying on embeddings or LLM inference during deployment. Our method operates in three key stages: (1) LLM-guided discovery of interpretable sub-intents and structured data generation, (2) pretraining a Non-Negated Tsetlin Machine (NTM) to extract semantic clause patterns, and (3) semantic enrichment of real data using these patterns, followed by fine-tuning a standard TM. The resulting pipeline remains fully interpretable, data-efficient, and domain-adaptable.

\subsection{LLM-Guided Sub-intent Discovery and Semantic Bootstrapping}
Large language models (LLMs) have demonstrated strong capabilities in extracting structured semantic information through prompting, instruction tuning, and zero-shot reasoning~\cite{Wang2023LargeLM,tripathi2025confidence}. We leverage these capabilities not for direct classification, but for generating symbolic scaffolds that guide clause discovery in symbolic models such as Tsetlin Machines (TMs). Rather than predicting labels, we prompt the LLM to identify \textit{fine-grained sub-intents}—interpretable drivers that explain why a sample belongs to a particular class.

To support generalization across domains, we design a generic prompt template that operates over any labeled dataset. The prompt includes placeholders for dataset name, domain description, and class labels, and instructs the model to generate sub-intents in the desired format. These sub-intents serve as symbolic anchors for interpretable clause learning and help inject domain knowledge into downstream models. For all synthetic data generation, we used \texttt{gpt-4o} (Azure OpenAI, May 2024 release). We adopted nucleus sampling with $p=0.9$, temperature = 0.7.

To simulate sub-intent distributions and build symbolic associations, we design a three-stage synthetic data generation pipeline rather than relying on a single prompt. This multi-step generation process is crucial for clause-based models like the TM, which require stability in lexical patterns to learn interpretable clauses, as well as exposure to syntactic and semantic diversity to avoid overfitting to surface form~\cite{Yadav2022RobustIT}. \textbf{Seed Stage.} The LLM is prompted with real domain-specific examples and asked to generate short, semantically faithful samples (approximately 15–20 words) for a given sub-intent. These serve as canonical expressions and provide clause-level anchors for the TM to begin learning consistent patterns.  \textbf{Core Stage.} Using the seed examples as anchors, the LLM generates structurally varied yet lexically stable samples. This variation ensures that the TM learns invariant features across syntax, which is essential for symbolic models that cannot rely on latent context to smooth over form variation~\cite{li-etal-2023-synthetic}. \textbf{Enriched Stage.} Finally, the LLM is prompted to introduce novel but semantically aligned expressions through modifiers, synonyms, and compositional phrasing. This stage expands the lexical space while maintaining intent consistency, helping TM clauses generalize beyond exact string matches.

This multi-stage strategy reflects the principles of curriculum learning~\cite{Bengio2009CurriculumL}, where simpler, canonical cases are learned first before introducing complexity. Prior work has shown that LLMs, when prompted with complex generation tasks in a single step, tend to collapse into high-probability patterns or overly generic phrasing~\cite{Yun2025ThePO}. By explicitly structuring generation into progressive stages, we ensure coverage, lexical variation, and semantic fidelity—each critical for effective clause formation in Boolean-symbolic models like the TM.

\subsection{Pretraining the Non-Negated Tsetlin Machine (NTM)}

To better understand our framework, we give an overview of the basic TM architecture to understand how the vanilla TM functions in general.

\subsubsection{Tsetlin Machine}

\begin{table}[t]
  \centering
  \caption{The Type I and Type II Feedback.}
  \label{table_combined}

  \resizebox{1\columnwidth}{!}{%
  \begin{tabular}{lcccccc}
    \hline
    \textbf{Input} & \multicolumn{3}{c}{\textbf{Type I Feedback}} & \multicolumn{3}{c}{\textbf{Type II Feedback}} \\
    \hline
     & \textbf{Clause} & \textbf{1} & \textbf{0} & \textbf{Clause} & \textbf{1} & \textbf{0} \\
     & \textbf{Literal} & \textbf{1 \ \ 0} & \textbf{1 \ \ 0} & \textbf{Literal} & \textbf{1 \ \ 0} & \textbf{1 \ \ 0} \\
    \hline
    \multirow{3}{*}{\textbf{Include Literal}} 
    & P(Reward)   & $\tfrac{s-1}{s}$ \ \ NA & 0 \ \ 0 & P(Reward)   & 0 \ \ NA & 0 \ \ 0 \\[1mm]
    & P(Inaction) & $\tfrac{1}{s}$ \ \ NA & $\tfrac{s-1}{s}$ \ \ $\tfrac{s-1}{s}$ & P(Inaction) & 1.0 \ \ NA & 1.0 \ \ 1.0 \\[1mm]
    & P(Penalty)  & 0 \ \ NA & $\tfrac{1}{s}$ \ \ $\tfrac{1}{s}$ & P(Penalty)  & 0 \ \ NA & 0 \ \ 0 \\[1mm]
    \hline
    \multirow{3}{*}{\textbf{Exclude Literal}} 
    & P(Reward)   & 0 \ \ $\tfrac{1}{s}$ & $\tfrac{1}{s}$ \ \ $\tfrac{1}{s}$ & P(Reward)   & 0 \ \ 0 & 0 \ \ 0 \\[1mm]
    & P(Inaction) & $\tfrac{1}{s}$ \ \ $\tfrac{s-1}{s}$ & $\tfrac{s-1}{s}$ \ \ $\tfrac{s-1}{s}$ & P(Inaction) & 1.0 \ \ 0 & 1.0 \ \ 1.0 \\[1mm]
    & P(Penalty)  & $\tfrac{s-1}{s}$ \ \ 0 & 0 \ \ 0 & P(Penalty)  & 0 \ \ 1.0 & 0 \ \ 0 \\[1mm]
    \hline
  \end{tabular}
  }
\end{table}

The Tsetlin Machine (TM) learns correlations between features and labels using propositional logic~\cite{Granmo2018TheTM}. A propositional logic formula in the TM, known as a clause, is a conjunction of negated and non-negated forms of the input features. These features are referred to as literals and are controlled by a set of Tsetlin Automata (TAs). Each input feature corresponds to two TAs: one (\textit{TA}) controls the original (non-negated) form of the literal, while the other (\textit{TA'}) controls its negation. Each TA decides whether to include or exclude the literal and operates with two actions (Include/Exclude) across $2N$ states. When a TA moves from state $1$ to $N$, it performs the Exclude action; when it moves from state $N+1$ to $2N$, it performs the Include action. Each move of a TA is triggered by feedback in the form of Reward, Penalty, or Inaction~\cite{Granmo2018TheTM}.

\par The training of the TM, as detailed in~\cite{Granmo2018TheTM}, involves learning the optimal actions through feedback mechanisms, specifically Type I and Type II feedback, as shown in Table~\ref{table_combined}. Consider a training example \((\boldsymbol{X}, y)\), where the input vector \(\boldsymbol{X}\) is a bag-of-words (BoW) representation and \(y\) is the ground-truth label. The predicted label is denoted by \(\hat{y}\). Multiple teams of TAs are responsible for TM learning, with each clause being assigned one TA per literal. Each TA operates in an environment defined by the training samples and the updating rule.

\par In the TM, the feedback mechanism is crucial for learning. Type I feedback occurs during true positive cases, when the clause output and the target agree. This feedback reinforces the inclusion of literals that correctly identify the positive class by rewarding the TAs that contributed to the correct classification—moving them away from the center states (toward the endpoints) and solidifying their actions. Conversely, Type II feedback occurs during false positive cases, when the clause output is positive but the target is negative. This feedback penalizes the inclusion of literals that incorrectly identify the class by moving the TAs toward the center states, weakening their actions and potentially switching their decisions.

\par The most important component of TM is the clause, which represents a certain sub-pattern among a particular set of patterns. This sub-pattern is in propositional AND-form, making it highly interpretable and suitable for logical understanding of the task. To illustrate, consider a bag-of-words input \(X=[x_1, \cdots, x_n]\), where \(x_k \in \{0,1\}\) indicates the presence or absence of a word in a sentence, and \(n\) represents the vocabulary size. Assuming \(\gamma\) classes, with each class requiring \(\alpha\) clauses, the model comprises \(\gamma \times \alpha\) clauses \(C_{\iota}^{\kappa}\), defined as:
\vspace{-1.5mm}
\begin{equation}\label{eqn2}
    C_{\iota}^{\kappa} = \left(\bigwedge \limits _{k \in I_{\iota}^{\kappa}}{x_k} \right) \wedge \left(\bigwedge \limits _{k \in \bar I_{\iota}^{\kappa}}{\neg x_k} \right),
\end{equation}

where \(I_{\iota}^{\kappa}\) and \(\bar I_{\iota}^{\kappa}\) are non-overlapping subsets of the input variable indices, \(I_{\kappa}^{\iota}, \bar{I_{\kappa}^{\iota}} \subseteq \{1, \cdots, n\}, I_{\kappa}^{\iota} \cap \bar{I_{\kappa}^{\iota}} = \emptyset\). The set \(I_{\iota}^{\kappa}\) includes indices of features that TAs include in their original form, while \(\bar I_{\iota}^{\kappa}\) includes indices of features in their negated form.

\par In each class, clauses with odd indexes are assigned positive polarity (+), voting in favor of the target class, while those with even indexes are assigned negative polarity (-), voting against it. The overall output for a class is obtained by subtracting the number of negative votes from the number of positive votes, as follows:
\vspace{-1mm}

\begin{equation}\label{eqn3}
    f^{\kappa}(X) = \sum_{{\iota}=1,3,\ldots}^{{\alpha}-1} C^{\kappa}_{\iota}{(X)} - \sum_{{\iota}=2,4,\ldots}^{\alpha} C^{\kappa}_{\iota}{(X)}.
\end{equation}

\par For \(\gamma\) classes, the final classification output \(\hat{y}\) is determined by the argmax operator, which selects the class with the highest net sum of votes:
\vspace{-1mm}
\begin{equation}\label{eqn4}
    \hat{y} = \mathrm{argmax}_{\kappa} \left( f^{\kappa}(X) \right).
\end{equation}

\subsection{Non-Negated Tsetlin Machine (NTM)}

To inject LLM-derived semantic knowledge into clause logic while preserving transparency, we introduce the Non-Negated Tsetlin Machine (NTM)—a variant of the standard TM tailored for symbolic semantic supervision. The NTM modifies the learning process in two key ways: (1) it eliminates negated literals from clause construction to improve monotonic interpretability, and (2) it applies \textit{boosted Type I feedback} to stabilize high-confidence literal selection.

\textbf{Clause Structure.}  
Unlike the standard TM which uses both positive and negated literals (Equation~\ref{eqn2}), the NTM restricts clause formation to only original (positive) literals. Each clause becomes a purely monotonic conjunction:
\begin{equation}
    C_{\iota}^{\kappa} = \bigwedge_{k \in I_{\iota}^{\kappa}} x_k,
\end{equation}
where $I_{\iota}^{\kappa} \subseteq \{1, \dots, n\}$ denotes the index set of included features. This modification reduces clause complexity and ensures all learned rules reflect affirmatively correlated lexical patterns, which aligns with the goal of identifying constructive sub-intent semantics.

\textbf{Feedback Adjustment.}  
To further enhance stability and interpretability, the NTM modifies the standard probabilistic Type I update by applying a boosted scheme. Specifically, for TAs associated with literals contributing to a correct prediction, the reward probability is increased from the standard $\frac{s-1}{s}$ to 1.0, and penalties are disabled (set to 0). Formally:

\begin{itemize}
    \item Type I Reward: $P_{\text{reward}} = 1.0$
    \item Type I Penalty: $P_{\text{penalty}} = 0.0$
\end{itemize}

This aggressive reinforcement encourages TAs to quickly converge on a high-confidence set of literals per clause that robustly define the sub-intent semantics from the synthetic dataset. Importantly, \textit{Type II feedback remains active} in the NTM, as in the standard TM, to suppress false positives and maintain class-level discrimination. Since NTM excludes negated literals by design, all feedback (including penalties) applies only to non-negated literals, ensuring semantic consistency and interpretability.

\textbf{Literal Extraction.}  
After pretraining the NTM on the synthetic samples generated for each sub-intent, we analyze the final states of the TAs associated with each clause. Literals with the deepest TA states (i.e., those furthest from the Exclude threshold) are selected as the most confident semantic indicators of that sub-intent. These literals serve as interpretable symbolic features and are retained for enriching real-world samples in the fine-tuning stage.

This simplification of clause structure and adjustment of feedback dynamics makes the NTM an ideal intermediate module for transferring LLM-inferred symbolic knowledge into clause-based learners without introducing inference-time complexity.

\subsection{Semantic Feature Injection and TM Fine-tuning}

The final stage of the framework injects the extracted semantic knowledge into real labeled data. Each sample in the original dataset is passed through the NTM, which outputs predicted sub-intents and their activated clauses. From these, we collect the corresponding high-confidence literals and append binary indicators of their presence to the sample's original bag-of-words (BoW) representation.

This enriched Boolean input combines direct lexical evidence with symbolic semantic abstractions derived from LLM-guided pretraining. A standard TM is then fine-tuned on this augmented dataset using both Type I and Type II feedback. During training, the TM learns interpretable clause logic over both raw and semantically inferred inputs.

Importantly, the entire pipeline introduces no new components at inference: the final model remains purely symbolic and efficient. All semantic augmentation occurs offline during training, and the prediction process uses clause voting based on Boolean inputs alone. This design preserves the transparency, auditability, and deployability of the TM while leveraging LLMs to guide symbolic generalization during learning.

\subsection{Interpretable Learning in NTM: Sub-Intent Clause Structuring}

The Non-Negated Tsetlin Machine (NTM) produces inherently interpretable models by learning symbolic clauses that capture coherent semantic patterns. Interpretability arises from two constraints: (i) exclusion of negated literals, yielding monotonic conjunctions, and (ii) boosted Type I feedback, which deterministically rewards literals that co-occur with correct predictions. These design choices ensure clauses are both human-readable and semantically meaningful.

Unlike prior work where clauses are trained on coarse class labels, our framework uses fine-grained \textit{sub-intents}—interpretable, LLM-derived factors such as \texttt{politics\_due\_to\_election}. Each sub-intent is assigned a dedicated pool of clauses (150 in our setup), allowing extraction of class-specific symbolic features without polarity assignments.

\paragraph{Monotonic Clause Semantics}
Each clause is a conjunction of included features:
\begin{equation}
C_j^\kappa(X) = \prod_{x_k \in I_j^\kappa} x_k,
\end{equation}
where $I_j^\kappa \subseteq \{1, \dots, n\}$ indexes features selected by Tsetlin Automata (TAs). Clauses activate only in the presence of affirmatively correlated features, ensuring transparent behavior.

\paragraph{Automaton-Guided Literal Selection}
Each literal $x_k$ is governed by a TA with state $\phi_{jk} \in \{1,\dots,2N\}$, determining whether it is included ($\phi_{jk} > N$) or excluded ($\phi_{jk} \leq N$). We define confidence as
\begin{equation}
\text{Conf}_{jk} = \max(0, \phi_{jk} - N),
\end{equation}
and select literals with $\text{Conf}_{jk}$ exceeding a stability threshold as semantic cues. We fixed the stability threshold at $\delta=5$ across all datasets, following standard TM heuristics.

\paragraph{Boosted Feedback Dynamics}
NTM modifies standard Type I feedback by setting $P_{\text{reward}}=1.0$ and $P_{\text{penalty}}=0.0$, ensuring stable inclusion of semantically important literals. Type II feedback remains unchanged to suppress false positives.

\paragraph{Clause Evolution Example}
On synthetic sub-intent samples, NTM clauses converge to compact symbolic rules. For instance:
\begin{itemize}
    \item $C^{\texttt{politics\_due\_to\_election}} = \texttt{parliament} \wedge \texttt{election}$
    \item $C^{\texttt{sports\_due\_to\_weather}} = \texttt{championship} \wedge \texttt{rain}$
\end{itemize}
These clauses evolve into sub-intent–specific literal clusters (e.g., \{\texttt{parliament}, \texttt{election}, \texttt{results}\}), which form symbolic embeddings used for downstream fine-tuning. Additional examples are included in Appendix B.

\section{Interpretable Feature Groups via Sub-Intents}
\label{sec:subintent_featuregroups}

To assess the semantic coherence and interpretability of symbolic learning guided by LLM-derived sub-intents, we extract representative feature groups from trained Non-Negated Tsetlin Machines (NTMs). Each sub-intent is associated with a dedicated set of clauses, and within these, we identify high-confidence literals based on Tsetlin Automaton (TA) states.

These literals—interpreted as semantic keywords—form a symbolic signature of the sub-intent. Concretely, for each clause \(\mathcal{C}_j^\kappa\) belonging to sub-intent \(\kappa\), we select the top-ranked literals \(\{x_k\}\) whose automaton states \(\phi_{jk} > N + \delta\) exceed a confidence threshold. The union of such literals across clauses yields the final feature group for sub-intent \(\kappa\).

This process yields symbolic representations that are:
\begin{itemize}
    \item statistically grounded, as they are reinforced by consistent Type I feedback across samples.
    \item semantically aligned, as sub-intents were derived from LLMs trained on rich world knowledge.
    \item interpretable, as all literals correspond to raw input vocabulary—no embeddings or opaque transformations are involved.
\end{itemize}

Below, we present qualitative examples of symbolic feature groups for selected sub-intents across multiple domains and datasets. These reflect what the model has learned to associate with specific semantic drivers.

\vspace{-1mm}
\subsection*{Hallmarks of Cancer (HoC)}
Sub-intent: \textit{activating invasion and metastasis\_due\_to: Activation of Notch1 signaling pathway under hypoxia} \\
Feature group: {cell, activation, invasion, signaling, carcinoma, metastasis, hypoxia, pathway}

Sub-intent: \textit{deregulating cellular energetics\_due\_to: ALA inhibiting Warburg effect and inducing cancer cell death} \\
Feature group: {ALA, Warburg, glycolysis, cell, apoptosis, cancer, metabolism, death}

Sub-intent: \textit{sustaining proliferative signaling\_due\_to: Overexpression of growth factor receptors} \\
Feature group: {receptor, proliferation, EGFR, growth, tumor, overexpression, activation, cancer}

\vspace{-1mm}
\subsection*{AG News (Topic Classification)}
Sub-intent: \textit{topic\_acquisition\_negotiations} \\
Feature group: {negotiation, merger, agreement, discussion, deal, acquisition, conflict, proposal}

Sub-intent: \textit{topic\_agriculture\_subsidies} \\
Feature group: {agriculture, farmer, subsidy, grant, aid, crop, finance, support}

\vspace{-1mm}
\subsection*{IMDb (Sentiment)}
Sub-intent: \textit{positive\_due\_to\_plot} \\
Feature group: {story, plot, engaging, intriguing, development, twist, script, writing}

Sub-intent: \textit{positive\_due\_to\_acting} \\
Feature group: {actor, performance, cast, believable, expressive, delivery, role, talent}

Sub-intent: \textit{negative\_due\_to\_plot} \\
Feature group: {predictable, slow, boring, weak, nonsense, messy, confusing, dull}

Sub-intent: \textit{negative\_due\_to\_acting} \\
Feature group: {overacting, flat, awkward, unconvincing, bland, wooden, poor, amateur}

\section{Experiments and Results}
\subsection{Datasets}

To evaluate the robustness and generalizability of our approach, we selected benchmark datasets across three domains: topic classification, sentiment analysis, and biomedical classification. \textbf{AG News}~\cite{Zhang2015CharacterlevelCN}: A news categorization dataset with four classes—World, Sports, Business, and Sci/Tech. \textbf{R8 \& R52}: Subsets of the Reuters 21578 corpus with 8 and 52 topic labels respectively, commonly used for symbolic text classification.  \textbf{IMDb}: A binary sentiment dataset of long-form movie reviews, capturing nuanced opinionated text. \textbf{SST-2}: A sentence-level sentiment classification dataset of short movie phrases with binary labels. \textbf{HoC}~\cite{baker2017cancer}: The Hallmarks of Cancer dataset with biomedical abstracts labeled for cancer-related biological processes.

\subsection{Experimental Setup}

We pretrained the Non-Negated Tsetlin Machine (NTM) using synthetic data generated via LLM-guided semantic bootstrapping. For each sub-intent, we generated \textbf{50 seed samples}, \textbf{50 core samples}, and \textbf{100 enriched samples}, totaling \textbf{200 samples per sub-intent}. This curriculum-style generation promotes lexical diversity while maintaining semantic consistency. For all synthetic data generation, we used gpt-4o accessed via the Azure OpenAI API. Prompts for sub-intent discovery and Seed/Core/Enriched generation are included in Appendix A. We used nucleus sampling with $p=0.9$, temperature = 0.7.

For NTM pretraining, we used $C = 150$ clauses per sub-intent to capture diverse monotone patterns. Unlike the standard TM, the NTM employs only positive literals, so polarity assignments (positive/negative clause pairs) are not used in this stage. The threshold was set to $T=5000$ to ensure sufficient evidence accumulation, and specificity to $s=5$ to balance inclusion probability and avoid overfitting. During fine-tuning, we enriched real labeled data with high-confidence literals extracted from NTM clauses and trained a standard TM with both positive and negated literals, again with $s=5$ for clause inclusivity.

To increase sensitivity to clause strength, we adopted the integer-weighted TM variant of~\cite{Abeyrathna2020ExtendingTT}. While our equations in Section~\ref{sec:proposed} describe the unweighted scoring function for clarity, in practice clause votes were weighted according to this scheme. This detail ensures consistency with prior work. 

We note that hyperparameter choices ($C$, $T$, $s$) were based on values commonly used in TM literature~\cite{Gorji2019ATM} and found empirically stable. However, we did not conduct full ablations (e.g., varying $s$, number of synthetic samples, or weighting scheme), which limits causal attribution of the observed gains. We leave a systematic ablation study to future work.

\begin{table*}[h]
  \caption{
    Performance of the proposed LLM-Guided TM on selected classification datasets. 
    The bold entries indicate the highest performance, while bold and underlined indicate the second highest. 
    A dash (--) indicates missing values not reported in literature. 
    LLM-Guided TM results are averaged over 5 runs, and the includes standard deviation. Baseline results are drawn from literature and do not report variances, so significance testing could not be applied. 
  }
  \label{tab:sota}
  \centering
  \scalebox{0.85}{
  \begin{tabular}{l|c|c|c|c|c|c}
    \toprule
    \textbf{Model/Datasets} & \textbf{AG-News} & \textbf{R8} & \textbf{R52} & \textbf{IMDB} & \textbf{SST2} & \textbf{HoC} \\
    \midrule
    BoW TFIDF  & 89.64  & 93.74 & 86.95 &  85.30 & 75.50 & 73.4 \\
    char-CNN & 87.20  & 94.02 & 85.37 & 80.35 &  80.78 & 76.60 \\
    fastText, h = 10  & 91.50 &   98.10 &  94.74 &  -- &  80.1 & --\\
    RNN/LSTM & 86.06  & 93.68 & 85.54 & 78.82 &  77.50 & 75.70 \\
    RNN/LSTM (GloVe) & 92.10  & 96.09 & 90.48 & 80.72 & 80.35 & 76.26 \\
    BERT & \textbf{94.75} & \underline{\textbf{97.49}} & \underline{\textbf{94.26}} & \textbf{93.46} & \textbf{94.00}  & \textbf{82.90} \\
    TM & 88.34 & 96.16 & 84.62 & 90.62 & 75.61 & 77.42 \\
    TM (GloVe) & 90.12 & 97.50 & 89.14 & 90.88 & 76.38 & 78.78 \\
    \midrule
    LLM-Guided TM & \underline{\textbf{93.10}} $\pm$ 0.96 & \textbf{97.88} $\pm$ 0.29 & \textbf{94.45} $\pm$ 0.33 & \underline{\textbf{92.10}} $\pm$ 0.68 & \underline{\textbf{85.24}} $\pm$ 1.12 & \underline{\textbf{81.90}} $\pm$ 1.40 \\

    \bottomrule
  \end{tabular}}
\end{table*}
\subsection{Performance Comparison}

\vspace{2mm}
\noindent\textbf{Benchmark Selection.} The baseline models reported in Table~\ref{tab:sota} are drawn from a curated set of prior works. Our selection reflects the evolution of text representation techniques for classification—ranging from traditional methods such as bag-of-words (BoW) and TF-IDF, to neural architectures like CNNs and RNNs, and embedding-based models such as fastText and BERT. We conclude with the Tsetlin Machine (TM) and its GloVe-enhanced variant, which mark a shift toward symbolic and interpretable learning. Results for comparison are obtained from the following benchmarks: \textbf{BoW, TF-IDF}~\cite{SprckJones2021ASI,baker-etal-2016-cancer,Ali2021EnhancingTP,Liu2018TaskorientedWE}, \textbf{CNN, RNN/LSTM, RNN/LSTM (GloVe)}~\cite{Zhang2015CharacterlevelCN,Hochreiter1997LongSM,dong-de-melo-2018-helping}, \textbf{fastText}~\cite{joulin-etal-2017-bag,Munikar2019FinegrainedSC}, \textbf{BERT-base, BERT-large}~\cite{peng-etal-2019-transfer,Sun2019HowTF,chen-miyake-2021-label}, \textbf{TM, TM (GloVe)}~\cite{yadav-etal-2021-enhancing,Kadhim2024ExploringSS,bhattarai-etal-2024-tsetlin}.

\subsection{Performance Analysis}

Our proposed \textbf{LLM-Guided TM} demonstrates consistent performance improvements across all datasets when compared to the vanilla Tsetlin Machine (TM) and even TM variants enriched with GloVe embeddings. Notably, our model achieves significant gains on low-context or sentiment-heavy datasets like \textbf{SST2} (+9.63\% over TM and +8.86\% over TM (GloVe)) and \textbf{IMDB} (+1.48\% over TM), where clause-level understanding benefits greatly from semantically enriched features.On topic classification benchmarks, our model surpasses or matches neural models in accuracy. For instance, on \textbf{AG-News}, it outperforms TM by +4.76\%, TM (GloVe) by approximately +3\%, and narrows the gap to BERT to just 1.65\%. On \textbf{R8}, it slightly outperforms TM (GloVe) and approaches BERT (97.88\% vs. 97.49\%). Similarly, on \textbf{R52}, our model surpasses TM by nearly 10\% and TM (GloVe) by over 5\%, while performing comparably to BERT (94.45\% vs. 94.26\%). In the biomedical domain (\textbf{HoC}), we report micro-F1 for HoC which often contains noisy or sparsely annotated samples, our model achieves 81.90\%, reducing the performance gap with BERT (82.90\%) while outperforming all TM variants and neural baselines such as RNN and char-CNN.

These results collectively illustrate that our LLM-Guided TM delivers competitive performance on par with transformer-based models—while preserving the full interpretability and symbolic reasoning capabilities of clause-based learning. Crucially, the improvement stems from our LLM-guided NTM’s ability to learn semantically coherent and class-specific feature sets through interpretable clauses. These symbolic features, once transferred to the original data, provide enriched input representations that enable the downstream TM to fine-tune on more informative, context-aware signals—resulting in both accuracy gains and traceable decision logic.

\section{Empirical Analysis: Symbolic Enrichment Enables Generalization}
\label{sec:symbolic_boost}

To examine how symbolic enrichment improves model performance, we analyze cases \textit{misclassified by vanilla TM} but \textit{correctly classified after enrichment}. Vanilla TM, operating on bag-of-words (BoW) inputs, discards rare tokens and cannot capture synonyms or compositional structure, since clauses activate only on exact conjunctions of observed literals. These limitations hinder generalization to unseen or lexically varied expressions. Our pipeline mitigates this by prompting an LLM to generate sub-intents and diverse synthetic examples across Seed, Core, and Enriched stages. A Non-Negated TM (NTM) trained on this corpus extracts high-confidence literals spanning a broader vocabulary. Injecting these literals into real data equips the final TM with semantically informed features that are otherwise inaccessible.

Importantly, enrichment features are injected deterministically from NTM clause activations, ensuring that no LLMs are queried at inference. This preserves reproducibility and makes the final pipeline fully symbolic. Injected features are binary indicators derived deterministically from NTM clause activations, meaning that rare or decomposed forms can be recovered (e.g., \texttt{immune}, \texttt{suppression} from \texttt{immunosuppression}) while preserving auditability.

We illustrate this process with three representative case studies.

\vspace{1mm}
\noindent\textbf{Case Study 1: IMDb Sentiment Misclassification.}  
\textit{Original sample:} “A crazy ride with a twisted ending.”  
\textit{Issue:} Rare words \texttt{crazy}, \texttt{twisted} are dropped in preprocessing; remaining terms (\texttt{ride}, \texttt{ending}) are too weak for clause activation, leading to misclassification.  
\textit{LLM Sub-intent:} \texttt{positive\_due\_to\_plot}  
\textit{NTM Feature Group:} \{\texttt{plot}, \texttt{twist}, \texttt{engaging}, \texttt{storyline}, \texttt{writing}\}  
\textit{Enriched input:} ride, ending, plot, twist, engaging, storyline, writing  
\textit{Interpretation:} Enrichment reintroduces semantically linked cues (\texttt{plot}, \texttt{twist}), activating clauses such as \texttt{plot} $\wedge$ \texttt{twist} $\wedge$ \texttt{ending}, and yielding the correct positive label with clause-level justification.

\vspace{1mm}
\noindent\textbf{Case Study 2: AG News – Topic Category Recovery.}  
\textit{Original sample:} “Talks between the two nations collapsed over unresolved objections.”  
\textit{Issue:} Only diffuse tokens (\texttt{talks}, \texttt{collapsed}, \texttt{nations}) remain, leading to misclassification as \texttt{World}.  
\textit{LLM Sub-intent:} \texttt{topic\_acquisition\_negotiations}  
\textit{NTM Feature Group:} \{\texttt{negotiation}, \texttt{agreement}, \texttt{deal}, \texttt{proposal}, \texttt{conflict}\}  
\textit{Enriched input:} talks, collapsed, nations, negotiation, agreement, deal, proposal, conflict  
\textit{Interpretation:} Negotiation-related features trigger business-specific clauses such as \texttt{proposal} $\wedge$ \texttt{conflict}, correcting the prediction and providing a transparent rationale.

\vspace{1mm}
\noindent\textbf{Case Study 3: HoC Biomedical Tagging.}  
\textit{Original sample:} “There was no evidence of immunosuppression.”  
\textit{Issue:} The compound \texttt{immunosuppression} is opaque to BoW, which lacks frequent base forms like \texttt{immune} and \texttt{suppression}.  
\textit{LLM Sub-intent:} \texttt{immune\_evasion\_due\_to\_suppression}  
\textit{NTM Feature Group:} \{\texttt{immune}, \texttt{suppression}, \texttt{evasion}, \texttt{resistance}, \texttt{modulation}\}  
\textit{Enriched input:} immunosuppression, immune, suppression, evasion, resistance, modulation  
\textit{Interpretation:} Decomposition introduces transparent triggers (\texttt{immune}, \texttt{suppression}, \texttt{evasion}), enabling clauses such as \texttt{immune} $\wedge$ \texttt{suppression} to fire, linking the sample to immune evasion. The result is a symbolically justified correction.

\vspace{1mm}
\noindent These analyses highlight the core contribution of symbolic enrichment: it recovers semantically aligned cues absent in surface tokens, enabling clauses to generalize beyond strict lexical overlap. Gains are strongest in domains with compositional or sparse vocabulary (e.g., biomedical text), while sentiment tasks remain more brittle, consistent with prior findings on the limits of synthetic augmentation~\cite{li-etal-2023-synthetic}.

\section{Conclusion}

We presented a novel framework that bridges the gap between large language models (LLMs) and interpretable clause-based learners by using LLMs to semantically bootstrap the Tsetlin Machine (TM). Our method introduces a three-stage generation process—seed, core, and enriched—that produces high-quality, semantically structured training data grounded in fine-grained sub-intents. A Non-Negated TM (NTM) trained on this data extracts symbolic, human-interpretable feature sets, which are then used to enrich real-world samples for fine-tuning a standard TM. Through extensive experiments across sentiment analysis, topic classification, and biomedical datasets, we demonstrate that our LLM-Guided TM consistently outperforms traditional TM variants and approaches the performance of state-of-the-art transformer models—all while maintaining full transparency and clause-level interpretability. Importantly, our approach requires no LLMs or embeddings at inference time, making it lightweight, transparent, and ideal for deployment in real-time and resource-constrained environments—offering a compelling solution where explainability, efficiency, and performance must coexist.

\section{Limitation}

Despite its promising results, our framework has several limitations. It depends on LLM-generated data, which may introduce subtle biases or hallucinated patterns into symbolic learning. The assumption that LLM-derived sub-intents accurately represent latent class semantics may not hold in complex or overlapping domains. In addition, removing negated literals in the Non-Negated TM improves interpretability but reduces expressive power, limiting the model’s ability to capture negation or contrastive logic. Finally, our evaluation focuses mainly on accuracy and qualitative interpretability, leaving robustness, clause stability, and hyperparameter sensitivity for future exploration.

\section{Acknowledgements}
This research is sponsored by funding from Stanford’s Center for Sustainable Development and Global Competitiveness (SDGC) and the Yonghua Foundation.

\bibliography{custom}

\clearpage

\section{Appendix A: Prompt Design for Sub-Intent Discovery and Sample Generation}
\label{appendix:prompts}

Our framework leverages structured prompt engineering to extract fine-grained sub-intents and generate domain-aligned training data using large language models (LLMs). This appendix describes the exact prompt formulations used for (1) sub-intent discovery, and (2) three-stage sample generation: seed, core, and enriched.

\subsection{A.1 Sub-Intent Discovery Prompt}

To identify fine-grained semantic sub-intents, we use a domain-aware zero-shot prompt that incorporates class metadata and instructs the LLM to infer concrete reasons for class membership.

\paragraph{Prompt Template:}
\scriptsize
\begin{quote}\ttfamily
You are an AI assistant tasked with analyzing biomedical texts from the [\{DATASET\_NAME\}] dataset. This dataset consists of medical research documents related to [\{DOMAIN\_DESCRIPTION\}], and each document is labeled with one or more of the following categories: [\{CLASS\_LABELS\}].

Your task is to identify fine-grained sub-intents for each class label. A sub-intent is a specific, grounded reason or topic that helps explain why a document belongs to a given category. Each sub-intent should be expressed using the format: classlabelname\_due\_to: explanation

Guidelines:
- Sub-intents must be specific, non-generic, and directly related to the context in the samples.
- Use terminology appropriate to the domain.
- Avoid duplicates and near-duplicates; merge similar expressions into a single, unified sub-intent.
- Do not include commentary or bullet points. Just return a list of sub-intents, one per line.

Output format:
classlabel\_due\_to: explanation \\
classlabel\_due\_to: explanation \\
...
\end{quote}
\normalsize

\paragraph{Example Output:}
\scriptsize
\begin{quote}\ttfamily
oncogenesis\_due\_to: gene expression linked to tumor suppressors \\
inflammation\_due\_to: elevated cytokine response following infection
\end{quote}
\normalsize

\subsection{A.2 Synthetic Sample Generation Prompts}

Sub-intents guide controlled sentence generation in three stages: seed, core, and enriched. Each prompt adapts to the input category and example set.

\subsubsection*{A.2.1 Seed Prompt}
\scriptsize
\begin{quote}\ttfamily
Category: \{CATEGORY\_LABEL\}

Below are real example sentences:
- \{Example 1\}
- \{Example 2\}

Now generate \{N\} new sentences that reflect this category.

Your new sentences should:
- Be \textbf{at least 15–20 words long}
- Introduce \textbf{new vocabulary} that is contextually consistent with the examples
- Use relevant terminology, synonyms, or modifiers that naturally fit the theme
- Preserve the logic and context of the category, while \textbf{extending its vocabulary coverage}

Output one sentence per line. Do not include comments or labels.
\end{quote}
\normalsize

\subsubsection*{A.2.2 Core Prompt}
\scriptsize
\begin{quote}\ttfamily
Category: \{CATEGORY\_LABEL\}

Here are example sentences for this category:
- \{Seed Example 1\}
- \{Seed Example 2\}

Generate \{N\} new sentences expressing the same theme.

Your sentences must:
- Be \textbf{at least 15–20 words long}
- Maintain the same context and meaning as the examples
- \textbf{Vary the vocabulary} by introducing new relevant terms, phrase structures, or expressions
- Expand the conceptual boundaries of the category while staying grounded in its core idea

Use formal and precise language. Output one sentence per line.
\end{quote}
\normalsize

\subsubsection*{A.2.3 Enriched Prompt}
\scriptsize
\begin{quote}\ttfamily
Category: \{CATEGORY\_LABEL\}

Here are several sentences that reflect this category using consistent language:
- \{Core Sample 1\}
- \{Core Sample 2\}

Now generate \{N\} new sentences that convey the \textbf{same theme}, but \textbf{expand the vocabulary} beyond what has already been used.

Your output should:
- Be \textbf{at least 15–20 words long}
- Use \textbf{semantically consistent but new terms} compared to the original examples
- Introduce \textbf{novel modifiers, synonyms, or phrasing} that enhance lexical diversity
- Avoid simple paraphrasing; instead, aim to \textbf{generalize or deepen} the expression through new vocabulary and structure

Output one sentence per line. No labels or commentary.
\end{quote}
\normalsize

\subsection{A.3 Prompt Design Objectives}

Across all stages, prompt design follows these core principles:

\begin{itemize}
    \item \textbf{Semantic precision}: Prompts elicit topic-specific, non-generic outputs.
    \item \textbf{Length and coverage}: All generated sentences meet minimum clause-level complexity and vocabulary extension.
    \item \textbf{Noise-free structure}: Output is clean, line-separated, and free of metadata or formatting noise.
\end{itemize}

These carefully controlled prompts enable LLMs to bootstrap symbolic models with rich, interpretable representations.

\section{Appendix B: Extended Clause Evolution Examples}
\label{sec:appendix}

Here we expand the Clause Evolution Example, showing the full set of sub-intents and literal clusters learned by the Non-Negated Tsetlin Machine (NTM).

\subsection*{Politics Sub-Intents}
\textbf{Examples:}  
\begin{itemize}
    \item \texttt{politics\_due\_to\_election}: ``The prime minister addressed the parliament after the election results.''  
    \item \texttt{politics\_due\_to\_debate}: ``A heated debate on foreign policy took place during the senate session.''  
\end{itemize}

\textbf{Learned Clauses:}  
\begin{itemize}
    \item $C_1^{\texttt{politics\_due\_to\_election}} = \texttt{parliament} \wedge \texttt{election}$
    \item $C_2^{\texttt{politics\_due\_to\_debate}} = \texttt{debate} \wedge \texttt{senate}$
\end{itemize}

\textbf{Semantic Clusters:}  
\begin{itemize}
    \item \texttt{politics\_due\_to\_election:} \{\texttt{parliament}, \texttt{election}, \texttt{minister}, \texttt{results}\}  
    \item \texttt{politics\_due\_to\_debate:} \{\texttt{debate}, \texttt{senate}, \texttt{policy}, \texttt{session}\}  
\end{itemize}

\subsection*{Sports Sub-Intents}
\textbf{Examples:}  
\begin{itemize}
    \item \texttt{sports\_due\_to\_performance}: ``Ronaldo scored twice but the rest of the match was uneventful.''  
    \item \texttt{sports\_due\_to\_weather}: ``The championship final was interrupted due to heavy rain.''  
\end{itemize}

\textbf{Learned Clauses:}  
\begin{itemize}
    \item $C_3^{\texttt{sports\_due\_to\_performance}} = \texttt{match} \wedge \texttt{scored}$
    \item $C_4^{\texttt{sports\_due\_to\_weather}} = \texttt{championship} \wedge \texttt{rain}$
\end{itemize}

\textbf{Semantic Clusters:}  
\begin{itemize}
    \item \texttt{sports\_due\_to\_performance:} \{\texttt{scored}, \texttt{goal}, \texttt{match}, \texttt{Ronaldo}\}  
    \item \texttt{sports\_due\_to\_weather:} \{\texttt{rain}, \texttt{stadium}, \texttt{interrupted}, \texttt{weather}\}  
\end{itemize}

\bigskip
These extended examples show how NTM clauses evolve from simple conjunctions into richer semantic clusters that align with interpretable sub-intents. In the main text we present only a subset (one politics, one sports) for brevity; here, we provide the full set to support reproducibility and interpretability analysis.

\end{document}